\documentclass[10pt,twocolumn,letterpaper]{article}

\def\eg{\emph{e.g}\onedot}

\def\ie{\emph{i.e}\onedot}

\usepackage[pagenumbers]{cvpr} % To force page numbers, e.g. for an arXiv version

\usepackage{makecell} 
\usepackage{xcolor}
\usepackage{pifont}

\usepackage{soul}% for underlines
\usepackage{xcolor,colortbl} % for cell colors
\usepackage{changepage,threeparttable} % for wide tables

\usepackage{booktabs,makecell,multirow}
\usepackage{rotating}          % provides \rotatebox
\usepackage[table]{xcolor}     % optional row shading

\usepackage[table]{xcolor} % [table] is needed if you also use colortbl
\usepackage{nicematrix}
\usepackage{graphicx} % for \rotatebox (though \Block has rotate)
\newcommand{\cmark}{\textcolor{green}{\ding{51}}} % ✓
\newcommand{\xmark}{\textcolor{red}{\ding{55}}}   % ✗

\newcommand{\Distill}{D3still}

\usepackage{adjustbox}

\definecolor{cvprblue}{rgb}{0.21,0.49,0.74}
\usepackage[pagebackref,breaklinks,colorlinks,allcolors=cvprblue]{hyperref}

\def\paperID{10028} % *** Enter the Paper ID here
\def\confName{CVPR}
\def\confYear{2026}

\title{AsymLoc: Towards Asymmetric Feature Matching for Efficient Visual Localization}

\author{Mohammad Omama\thanks{Work done as a part of a summer internship at Amazon.}\\
The University of Texas at Austin\\
{\tt\small mohd.omama@utexas.edu}
\and
Gabriele Berton  \quad Eric Foxlin \quad Yelin Kim\\
Amazon\\
{\tt\small \{gberton, efoxlin, kimyelin\}@amazon.com}
}

\begin{document}
\maketitle

\begin{abstract}

Precise and real-time visual localization is critical for applications like AR/VR and robotics, especially on resource-constrained edge devices such as smart glasses, where battery life and heat dissipation can be a primary concerns. While many efficient models exist, further reducing compute without sacrificing accuracy is essential for practical deployment. To address this, we propose asymmetric visual localization: a large Teacher model processes pre-mapped database images offline, while a lightweight Student model processes the query image online. This creates a challenge in matching features from two different models without resorting to heavy, learned matchers.

We introduce AsymLoc, a novel distillation framework that aligns a Student to its Teacher through a combination of a geometry-driven matching objective and a joint detector-descriptor distillation objective, enabling fast, parameter-less nearest-neighbor matching.
Extensive experiments on HPatches, ScanNet, IMC2022, and Aachen show that AsymLoc achieves up to $\mathbf{95\%}$ of the teacher's localization accuracy using an order of magnitude smaller models, significantly outperforming existing baselines and establishing a new state-of-the-art efficiency-accuracy trade-off.

\end{abstract}
    
\section{Introduction}
\label{sec:intro}

\begin{figure}[t]
    \centering
    \includegraphics[width=\linewidth]{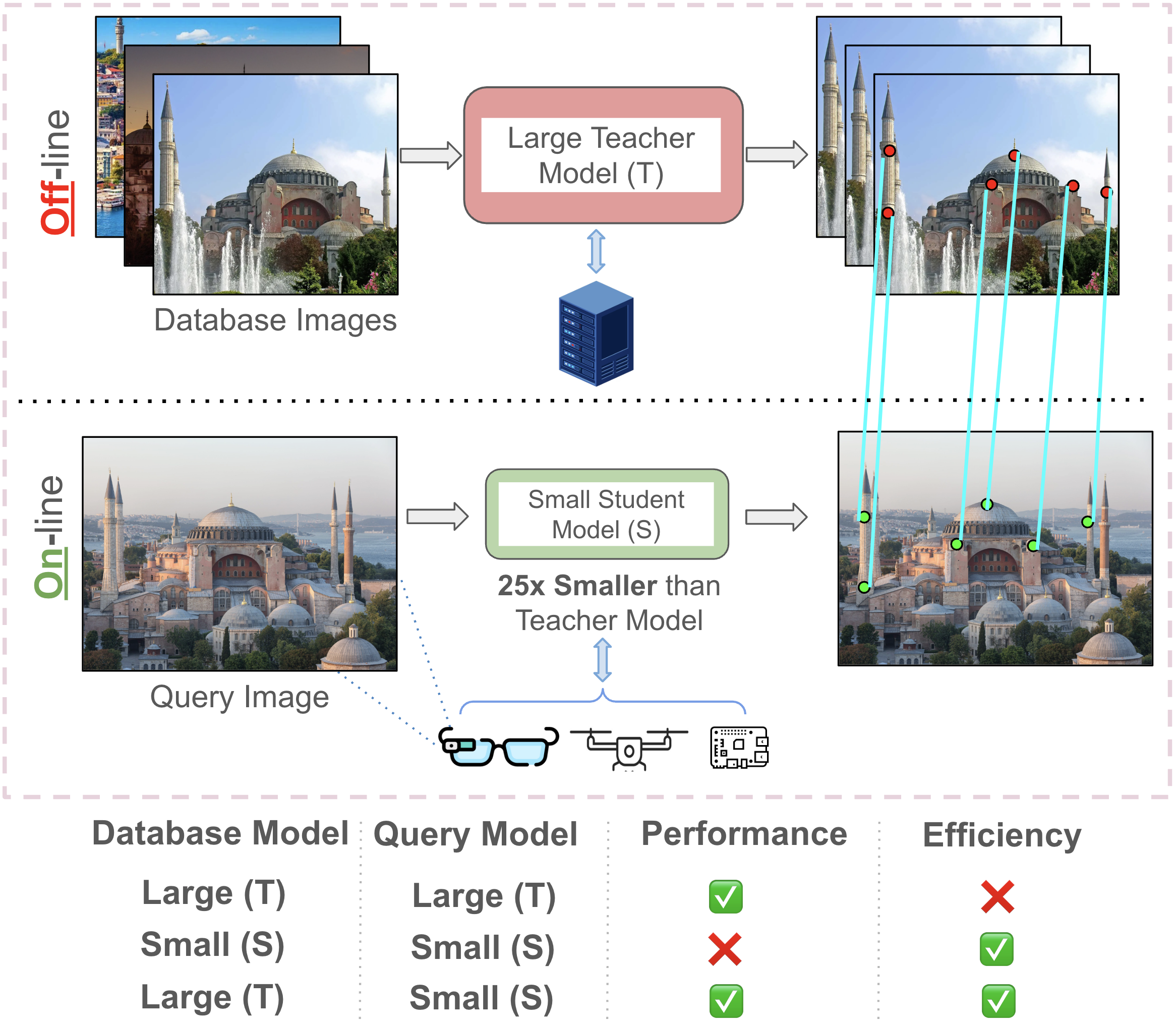}
    \vspace{-4mm}
    \caption{\textbf{AsymLoc bridges the gap between powerful database models and lightweight on-device localization.}  
    By explicitly modeling teacher–student asymmetry, AsymLoc enables compact query models to perform real-time localization on edge platforms such as \textbf{smart glasses, drones, and single-board computers}, while larger teacher models process the pre-mapped database images offline.  
    This design delivers near-teacher accuracy with up to $\mathbf{25\times}$ smaller models and a fraction of the compute cost.
    }
    \vspace{-4mm}
    \label{fig:teaser}
\end{figure}

% \gb{TODO we should cite AMES, they do re-ranking with asymmetric local features \url{https://arxiv.org/abs/2408.03282}}

Visual localization, the process of estimating a precise 6-DoF (degree of freedom) camera pose from a pre-mapped image database using only visual input \cite{sattler2018benchmarking}, is fundamental for applications like augmented reality (AR/VR) \cite{sarlin2022lamar} and robotics \cite{Blum_2025_crocodl}. These applications critically depend on obtaining precise pose estimates in real-time, often on resource-constrained edge devices. A typical pipeline \cite{Sarlin2018LeveragingDV, sarlin2019coarse} first selects a subset of neighboring (or similar) database images, often using GPS prior or visual place recognition (VPR), and then performs feature matching between the query and this subset.
The efficiency of this matching step is crucial, especially on edge devices such as smart glasses, where computation is limited by practical factors such as battery life and heat dissipation.

One common solution to improve deployment-time efficiency is to employ smaller models, a focus of many previous works \cite{howard2017mobilenets, sandler2018mobilenetv2}.
While smaller models naturally lead to cheaper computation, they can suffer from a non-negligible drop in accuracy \cite{tan2019efficientnet, zhai2022scaling}.
In this paper, we aim to build a new localization pipeline that approaches the accuracy of larger models while retaining the efficiency of smaller ones.

To this end, we leverage the insight that database images can be pre-processed offline, where computational constraints are not a concern. We therefore propose an \textbf{asymmetric visual localization} scenario: we use a large, high-performance \textit{Teacher} model for offline feature extraction on the database, and a small, efficient \textit{Student} model for online feature extraction on queries. While this naturally leads to faster computation, it raises the challenge of how to match features that are extracted from two different models.

While a solution to bridge this gap is to use learned matchers as SuperGlue \cite{superglue} or LightGlue \cite{lightglue2023},
this can be impractical in constrained devices (\eg LightGlue has over 10 times more parameters than common features extractors like SuperPoint \cite{superpoint}): we therefore aim to make the Teacher and Student features directly compatible with distillation, enabling the matching step to be performed with simple, fast, and parameter-less mutual nearest neighbor matching.

To this end we propose \textbf{AsymLoc}, a technique that aligns the representations of a small Student model to those of a frozen Teacher model. AsymLoc builds on the insight that alignment should occur in the \textit{joint detector–descriptor space}, where detection confidence modulates descriptor similarity. It achieves this by combining a geometry-driven matching objective with a probabilistic distillation loss that transfers the teacher’s joint matchability distribution to the student. This formulation couples detection and description supervision into a single differentiable objective, ensuring that student features remain natively compatible with teacher-derived map features.
To assess the robustness of AsymLoc, we perform a thorough experimental evaluation on a wide combination of multi-domain datasets (indoor, outdoor, cross-domain), multiple model sizes (with students up to 25 times smaller than the teacher), and teacher architectures (SuperPoint and SiLK).
Our results highlight the robustness of AsymLoc, which consistently outperform existing techniques, and achieves near-teacher localization accuracy at an order of magnitude lower compute, paving the way for very lightweight yet powerful visual localization pipelines.
An outline of AsymLoc is depicted in \Cref{fig:teaser}, which depicts how using such asymmetric setup can lead to good results and high efficiency.

\vspace{-1em}
\paragraph{Contributions.} Our main contributions are as follows:
\begin{enumerate}
    \item Driven by real-world constraints, we introduce the task of asymmetric visual localization, where a larger model is used to process map images, while a lightweight model is used on queries.
    \item We propose a novel \emph{joint detector–descriptor distillation} framework, called AsymLoc, that integrates detector confidence and descriptor similarity into a unified probabilistic alignment objective, coupled with a geometric matching loss.
    \item Thorough experiments show that AsymLoc consistently outperforms existing alternatives at the same inference cost, achieving 95.5\% (over SiLK) and 93\% (over SuperPoint) the accuracy of standard localization pipelines at an order of magnitude less inference cost on the popular Aachen dataset.
\end{enumerate}

\section{Related Work}
\label{sec:related}

\textbf{Visual Localization.}  6-DoF visual localization is primarily divided into structure-based and image-based methods.  \emph{Structure-based methods} perform direct 2D-to-3D matching, comparing keypoints from a query image against a 3D Structure-from-Motion (SfM) model generated using database images \cite{svarm2017city, toft2018semantic, sattler2017efficient, liu2017efficient, taira2018inloc}. While capable of precise poses, constructing (and extending) large-scale 3D models is still a significant challenge \cite{Sattler_2017_large_scale_3d}.

On the other hand
\emph{image-based methods}  
% 2D image retrieval-based methods 
only require a database of geo-tagged images, which is trivial to construct and to maintain.
Common image-based pipelines \cite{Sattler_2017_large_scale_3d, sattler2018aachen, hloc} rely on a two-step process: an image-retrieval-based search to get a shortlist of images to match to, performed with visual place recognition models \cite{NetVlad, CosPlace, Izquierdo_CVPR_2024_SALAD, MegaLoc}; and a second step consisting on image matching.
% rely on image retrieval \cite{NetVlad, torii201524, CosPlace, EigenPlaces, MegaLoc}, offering greater robustness by using global, image-wide information. However, their precision is limited, providing only an approximate pose based on the database's discretization, which is often insufficient \cite{sattler2018benchmarking, taira2018inloc}. Modern hierarchical localization \cite{hloc} first uses coarse retrieval to prune the search space, followed by 2D-3D descriptor matching.
While the asymmetric setting has been thoroughly explored for image retrieval \cite{Asym_distill_xie2024d3still, Asym_CSD_wu2022contextual, Asym_AML_budnik2021asymmetric, Asym_Hetroduggal2021compatibility, Asym_backward_shen2020towards, Asym_trans_hu2019towards},
no previous methods has explored the possibility of applying an asymmetric framework on the image matching step, making our work the first to tackle this important problem.

% These methods inherently assume a symmetric setting, meaning the same feature extractor is used for both database and query images. The contribution of AsymLoc is particularly significant because it directly relaxes this constraint. 

\textbf{Learned Detectors and Descriptors.}
Learned local features \cite{superpoint, yi2016lift, d2net, mishchuk2017hardnet, tian2017l2, luo2019contextdesc, tian2019sosnet, r2d2, tyszkiewicz2020disk, gleize2023silk} have significantly advanced feature matching in recent years.
% Notable works include SuperPoint \cite{superpoint}, which employs a self-supervised strategy using synthetic data and homographies. R2D2 \cite{r2d2} jointly learns detection, description, and a reliability score. DISK \cite{tyszkiewicz2020disk} uses reinforcement learning to optimize for correct matches. More recently, SiLK \cite{gleize2023silk} has shown that keypoints and descriptors can be trained on a large-scale, homography-adapted dataset with simple assumptions and losses, and it outperforms previous complex approaches. While learned features significantly improve feature matching, they assume a symmetric deployment and do not address compatibility between heterogeneous models.
Notable works include SuperPoint \cite{superpoint}, which adopts a self-supervised strategy based on synthetic data and homographies; D2Net \cite{d2net}, which learns dense, jointly invariant detection and description from image pairs;  DISK \cite{tyszkiewicz2020disk}, which leverages reinforcement learning to optimize for correct matches; and ALIKE \cite{Zhao2022ALIKE}, which focuses on lightweight, real-time local features suitable for deployment on resource-constrained devices. More recently, SiLK \cite{gleize2023silk} demonstrated that keypoints and descriptors can be effectively trained on a large-scale, homography-adapted dataset using simple assumptions and loss functions, outperforming more complex prior approaches. Learned detectors and descriptors  assume symmetric deployment and do not address compatibility across heterogeneous models.

\begin{figure*}[t]
    \centering
    \includegraphics[width=\linewidth]{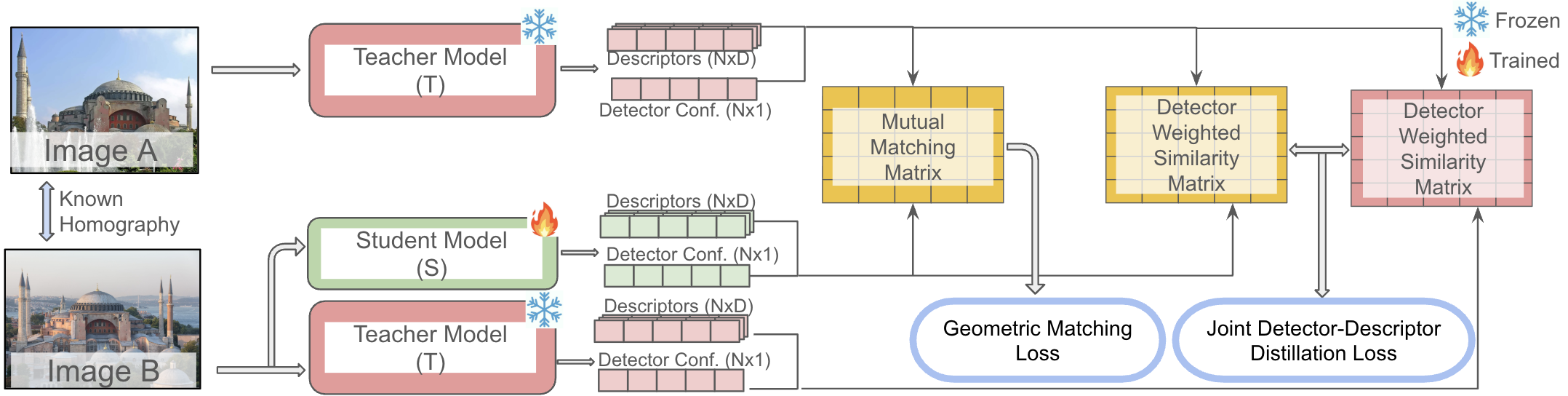}
    \caption{\textbf{AsymLoc Training Pipeline.} 
    Given a pair of images $(A, B)$ with known homography, the teacher model $T$ processes image $A$, while image $B$ is processed by both the teacher $T$ and the student $S$. 
    Each network produces $N$ keypoints with corresponding detector confidence and descriptors. 
    The teacher outputs from $A$ and the student outputs from $B$ are combined to form the \emph{Mutual Matching Matrix} (Sec.~\ref{sec:method_matching}), which is used to compute the geometric matching loss. 
    In parallel, we construct two detector-weighted similarity matrices: one with the teacher outputs of $A$ and the student outputs of $B$, and the other with the teacher outputs of $A$ and the teacher outputs of $B$. 
    These matrices form two joint detector–descriptor similarity spaces (Sec.~\ref{sec:method_distill}); their distributions are then aligned through a distillation loss.
    \vspace{-1em}
    }
    \label{fig:pipeline}
\end{figure*}

\textbf{Matchers.}  
Learned matchers such as SuperGlue~\cite{superglue}, SGMNet \cite{seededgraphmatching}, LightGlue~\cite{lightglue2023}, and OmniGlue \cite{omniglue}, 
are networks built on top of existing detectors and descriptors extractors to improve over standard mutual nearest neighbor matching.
These often rely on powerful graph neural network or transformer-based architectures, which use global information from both images to robustly match keypoints between two images.
Although effective, these methods require additional network components that add significant runtime and parameter overhead, vastly exceeding the size of the feature extractor itself.
For instance, SuperPoint contains roughly $1.3$M parameters and runs in under $10$ms per image pair, whereas LightGlue adds $\sim$13M parameters and increases inference time to about $93$ms on similar hardware \cite{Berton_2024_EarthMatch}.
While this might not be a problem in many robotics applications,
% that require visual localization, 
it can inflict a heavy toll on resource-constrained edge devices, such as smart glasses, where computation is limited and increasing battery life is crucial.
% Matchers enable cross-model alignment but are too heavy for edge devices like AR headsets. AsymLoc instead trains lightweight features to be \emph{natively compatible}, avoiding inference-time matcher overhead.

\textbf{Dense Methods.}
Dense (or semi-dense), end-to-end matching methods, such as LoFTR \cite{loftr}, RoMa \cite{densse_roma}, and others \cite{dense_minima_roma, dense_se2_loftr, efloftr, dense_aspanformer, dense_matchformer, dense_jung2025edm, Chen_2025_RDD}, process two images jointly in a single network to directly output matches. While achieving good results and being robust to large viewpoint changes, these methods typically have a large parameter count, making them unsuitable for resource-constrained edge devices. Moreover, because they require both images at inference time, they preclude the pre-computation of map descriptors and are not suitable for asymmetric settings.

\vspace{-1em}

\paragraph{Distillation.}
Knowledge distillation (KD) transfers generalization ability from a larger \emph{teacher} to a compact \emph{student} by matching softened output distributions  produced with a temperature-scaled softmax~\cite{hinton2015distilling}. This simple cross-entropy on soft targets regularizes the student beyond one-hot labels and has inspired a large body of follow-ups that enrich the supervision signal. Representative directions include deeper supervision via intermediate hints (FitNets)~\cite{romero2015fitnets}, attention map transfer~\cite{zagoruyko2017paying}, flow-of-solution-procedure (FSP) relations~\cite{yim2017gift}, variational information distillation (VID)~\cite{ahn2019variational}, among many others.
Beyond logits, \emph{feature distillation} ~\cite{romero2015fitnets,zagoruyko2017paying,yim2017gift,park2019relational,ahn2019variational,tian2020crd} aligns representations of the teacher and  the student to guide the student toward the teacher’s embedding geometry. These methods typically minimize the distance between the output features of the student network and the teacher network to guide the student network to generate similar features to those of the teacher network.

\paragraph{Asymmetry in image retrieval.}
% Asymmetry has been actively explored in global image retrieval. Hu et al.\ introduce \emph{feature translation} to bridge heterogeneous representations across models~\cite{hu2019featuretranslation}. Shen et al.\ propose \emph{backward-compatible training} to enable upgrading encoders without re-indexing galleries~\cite{shen2020bct}. Duggal et al.\ study \emph{compatibility-aware heterogeneous visual search} across mixed-feature databases~\cite{duggal2021compatibility}. Budnik and Avrithis formalize \emph{asymmetric metric learning (AML)} for teacher–student retrieval with a contrastive-style objective tailored to cross-model compatibility~\cite{budnik2021aml}. More recently, \emph{contextual similarity distillation (CSD)} transfers pairwise similarity structure rather than raw features, improving compatibility under capacity gaps~\cite{wu2022csd}. State-of-the-art \emph{D3Still} further emphasizes ranking-order consistency by distilling similarity \emph{differentials}, achieving superior cross-model alignment~\cite{xie2024d3still}.

Asymmetry has been actively explored in global image retrieval \cite{hu2019featuretranslation, shen2020bct, duggal2021compatibility, suma2024ames, budnik2021aml, wu2022csd}. \cite{hu2019featuretranslation} introduced \emph{feature translation} to bridge heterogeneous representations across models for image retrieval. \emph{Backward-compatible training} was introduced in \cite{shen2020bct} to enable upgrading encoders without re-indexing galleries. \emph{Compatibility-aware heterogeneous visual search} \cite{duggal2021compatibility} trains a large gallery model to be compatible with a small query model. \emph{Asymmetric metric learning (AML)} \cite{budnik2021aml} introduced an asymmetric distillation strategy for small retrieval models. \emph{contextual similarity distillation (CSD)}~\cite{wu2022csd} transfers pairwise similarity structure rather than raw features,
improving compatibility under capacity gaps. State-of-the-art \emph{D3Still}~\cite{xie2024d3still} further emphasizes ranking-order consistency by distilling similarity \emph{differentials}.

All of these methods focus exclusively on global descriptors. In contrast, we address the local detector–descriptor pipeline, where both \emph{where} to look (detectors) and \emph{how} to match (descriptors) must remain compatible across asymmetric teacher–student models.

\section{Methodology}
\label{sec:method}

Our goal is to design a visual localization pipeline that unlocks the efficiency of tiny models while retaining the accuracy of larger models.
To this end, we propose \textbf{AsymLoc}, the first visual localization framework made of two separate models: a larger teacher model, which processes the database images offline, and a small student model, which runs online and produces outputs that are consistent with those of the teacher.
The key insight is that compatibility should be learned through both geometric and probabilistic supervision: a geometric matching objective enforces spatial correspondence, while a joint detector–descriptor distillation loss ensures consistent feature interaction across models. We next formalize the problem in \Cref{sec:method_formulation}, and describe the two core objectives in \Cref{sec:method_matching} and \Cref{sec:method_distill}.

\subsection{Problem Formulation}\label{sec:method_formulation}

Let $\mathcal{I}_d$ denote a database image and $\mathcal{I}_q$ a query image.  
We consider two models:
\begin{itemize}
    \item A \emph{teacher} model $T$, a powerful network used offline to process database images.
    \item A \emph{student} model $S$, a lightweight network deployed online to process query images on-device.
\end{itemize}

\paragraph{Teacher features.}  
Applying $T$ to $\mathcal{I}_d$ yields a set of keypoints (detectors) and associated descriptors:
% \[
% \mathcal{F}_{m^T} = \{ (\mathbf{p}_i^T, \mathbf{d}_i^T) \}_{i=1}^{N},
% \]
% \[
% \{ (\mathbf{w}_i^T, \mathbf{d}_i^T) \}_{i=1}^{N} = T(\mathcal{I}_d),
% \]

{\small
\begin{equation}
   \{ (\mathbf{w}_i^T, \mathbf{d}_i^T) \}_{i=1}^{N} = T(\mathcal{I}_d),
\end{equation}
}

where $\mathbf{w}_i^T \in (0,1)$ denotes the detector confidence of the $i$-th keypoint and $\mathbf{d}_i^T \in \mathbb{R}^D$ its descriptor.

\paragraph{Student features.}  
Likewise, applying $S$ to $\mathcal{I}_q$ yields
% \[
% \mathcal{F}_{q^S} = \{ (\mathbf{p}_j^S, \mathbf{d}_j^S) \}_{j=1}^{N},
% \]
% \[
%  \{ (\mathbf{w}_j^S, \mathbf{d}_j^S) \}_{j=1}^{N} = S(\mathcal{I}_q),
% \]
{\small
\begin{equation}
   \{ (\mathbf{w}_j^S, \mathbf{d}_j^S) \}_{j=1}^{N} = S(\mathcal{I}_q),
\end{equation}
}\\
with $\mathbf{w}_j^S$ detector confidence and $\mathbf{d}_j^S$ its descriptors.

\paragraph{Pose estimation.}
In an asymmetric scenario, these feature are used to compute the matches, from which we can estimate the relative pose of a query image with respect to the database image:
% \[
% \mathbf{T}_{S(I_q)\rightarrow T(I_d)} \in SE(3).
% \]
{\small
\begin{equation}
    \mathbf{T}_{S(I_q)\rightarrow T(I_d)} \in SE(3).
\end{equation}
}
In a symmetric scenario, both query and map images are processed by the teacher $T$, and we obtain the reference transformation
% \[
% \mathbf{T}_{T(I_q)\rightarrow T(I_d)} \in SE(3).
% \]
{\small
\begin{equation}
    \mathbf{T}_{T(I_q)\rightarrow T(I_d)} \in SE(3).
\end{equation}
}

We want to ensure that the transformation estimated in the asymmetric case, $\mathbf{T}_{S(I_q)\rightarrow T(I_d)}$, closely approximates the one estimated in the symmetric case $\mathbf{T}_{T(I_q)\rightarrow T(I_d)} \in SE(3)$, obtained when both images are processed by the teacher. This would guarantee that the features extracted by the student are compatible with those extracted by the teacher, a key ingredient for asymmetric localization.
While the most straightforward way to achieve this is to naively apply distillation (\ie feed an image to both networks, and maximize the similarity of their outputs), we empirically find that this leads to unsatisfactory results (see \Cref{sec:experiments}), which is in line with similar findings in the asymmetric retrieval literature \cite{duggal2021compatibility}.
% , likely due to the fact that the standard training objectives of SILK \cite{gleize2023silk} and SuperPoint \cite{superpoint} already have a contrastive component which is a much stronger signal. We empirically show this in the results section. \mo{Take a look} 
Therefore, we instead propose to align student's outputs to the teacher's (both detector and descriptors outputs) by relying on a dataset of image pairs related by known homographies, following the process depicted in \Cref{fig:pipeline}.
% Following SILK~\cite{gleize2023silk}, we rely on image pairs related by known homographies, a setup shown to outperform R2D2~\cite{r2d2} and SuperPoint~\cite{superpoint}.
For each image pair, we define two complementary objectives: a \emph{geometric matching} loss and a novel \emph{joint detector–descriptor distillation} loss.
We describe these objectives in detail below.

\subsection{Geometric Matching Loss} \label{sec:method_matching}

The first objective of AsymLoc enforces geometric consistency between teacher–student feature pairs through a correspondence-based loss function. Rather than regressing descriptors directly, we operate at the level of \emph{probabilistic matches}, where both detector scores and descriptor similarities contribute to soft assignments.
To obtain these soft assignments we first introduce the concept of similarity matrix between two images: given two images $a$ and $b$ with a known homography relating their viewpoints, we extract local descriptors from the teacher model on image $a$, $\{\mathbf{d}_i^T(a)\}_{i=1}^{N}$, and from the student model on image $b$, $\{\mathbf{d}_j^S(b)\}_{j=1}^{N}$.  
We then compute a pairwise descriptor similarity matrix
{\small
\begin{equation}
   \mathbf{S}_{ij}^{TS} = \frac{\langle \mathbf{d}_i^T(a), \mathbf{d}_j^S(b) \rangle}{\tau},
\end{equation}
}
% \[
% \mathbf{S}_{ij}^{TS} = \frac{\langle \mathbf{d}_i^T(a), \mathbf{d}_j^S(b) \rangle}{\tau},
% \]
where $\langle \cdot , \cdot \rangle$ denotes the dot product and $\tau$ is a temperature parameter that controls the sharpness of similarity values. The superscript (TS) in $\mathbf{S}_{ij}^{TS}$ means that the first image was processed by teacher and the second by student.

\paragraph{Mutual matching matrix.}  
Given the teacher detector confidence $\mathbf{w}_i^T(a)$ of keypoint $i$ in image $a$, and $\mathbf{w}_j^S(b)$ the studnet detector confidence of keypoint $j$
in image $b$. We define the bi-directional, mutual matching matrix as:
{\small
\begin{equation}
   P_{ij}^{TS}
= \mathbf{w}_i^T(a)\, \mathbf{w}_j^S(b)\, 
\sigma_r(\mathbf{S}_{ij}^{TS})_{ij}\, \sigma_c(\mathbf{S}_{ij}^{TS})_{ij},
\end{equation}
}
% \[
% P_{ij}^{TS}
% = \mathbf{w}_i^T(a)\, \mathbf{w}_j^S(b)\, 
% \sigma_r(\mathbf{S}_{ij}^{TS})_{ij}\, \sigma_c(\mathbf{S}_{ij}^{TS})_{ij},
% \]
where $\sigma_r(\cdot)$ and $\sigma_c(\cdot)$ denote row- and column-wise softmax
normalizations, respectively:

{\small
\begin{equation}
\sigma_r({\mathbf{S}^{TS}})_{ij} = 
\frac{\exp({\mathbf{S}^{TS}}_{ij})}{\sum_k \exp({\mathbf{S}_{ik}^{TS}})},
\end{equation}
}
{\small
\begin{equation}
\sigma_c({\mathbf{S}^{TS}})_{ij} = 
\frac{\exp({\mathbf{S}^{TS}}_{ij})}{\sum_k \exp({\mathbf{S}_{kj}^{TS}})}.
\end{equation}
} \\
% \[
% \sigma_r({\mathbf{S}^{TS}})_{ij} = 
% \frac{\exp({\mathbf{S}^{TS}}_{ij})}{\sum_k \exp({\mathbf{S}_{ik}^{TS}})},
%  \]
%  \[
% \sigma_c({\mathbf{S}^{TS}})_{ij} = 
% \frac{\exp({\mathbf{S}^{TS}}_{ij})}{\sum_k \exp({\mathbf{S}_{kj}^{TS}})}.
% \]
This  yields a soft, detector-aware matching matrix,
ensuring that reliable keypoints dominate the correspondence distribution.
The same construction applies to $P_{ij}^{ST}$.

\paragraph{Geometric matching loss.}  
Given ground-truth correspondences $\mathcal{M}_{ab}$ derived from a known homography or epipolar geometry between images $a$ and $b$, we define the geometric matching loss as:

{\small
\begin{equation}
\mathcal{L}_{\text{match}}
= - \!\!\sum_{\substack{(i,j) \in \mathcal{M}_{ab} \\ \mathbf{w}_i^T(a) > \tau_d}}
  \log P_{ij}^{TS}
  - \!\!\sum_{\substack{(i,j) \in \mathcal{M}_{ab} \\ \mathbf{w}_i^T(b) > \tau_d}}
  \log P_{ij}^{ST},
\end{equation}
}
% \[
% \mathcal{L}_{\text{match}}
% = - \!\!\sum_{\substack{(i,j) \in \mathcal{M}_{ab} \\ \mathbf{w}_i^T(a) > \tau_d}}
%   \log P_{ij}^{TS}
%   - \!\!\sum_{\substack{(i,j) \in \mathcal{M}_{ab} \\ \mathbf{w}_i^T(b) > \tau_d}}
%   \log P_{ij}^{ST},
% \]
where $\tau_d$ is a confidence threshold applied to the \emph{teacher’s} detector confidence.  
The loss is computed only for keypoints that the teacher identifies as reliable (\ie, $\mathbf{w}_i^T > \tau_d$), ensuring that supervision originates from confident detections.  
This focuses learning on high-quality correspondences while avoiding the noise introduced by uncertain or low-confidence teacher keypoints.

\begin{figure}[t]
    \centering
    \includegraphics[width=0.85\linewidth]{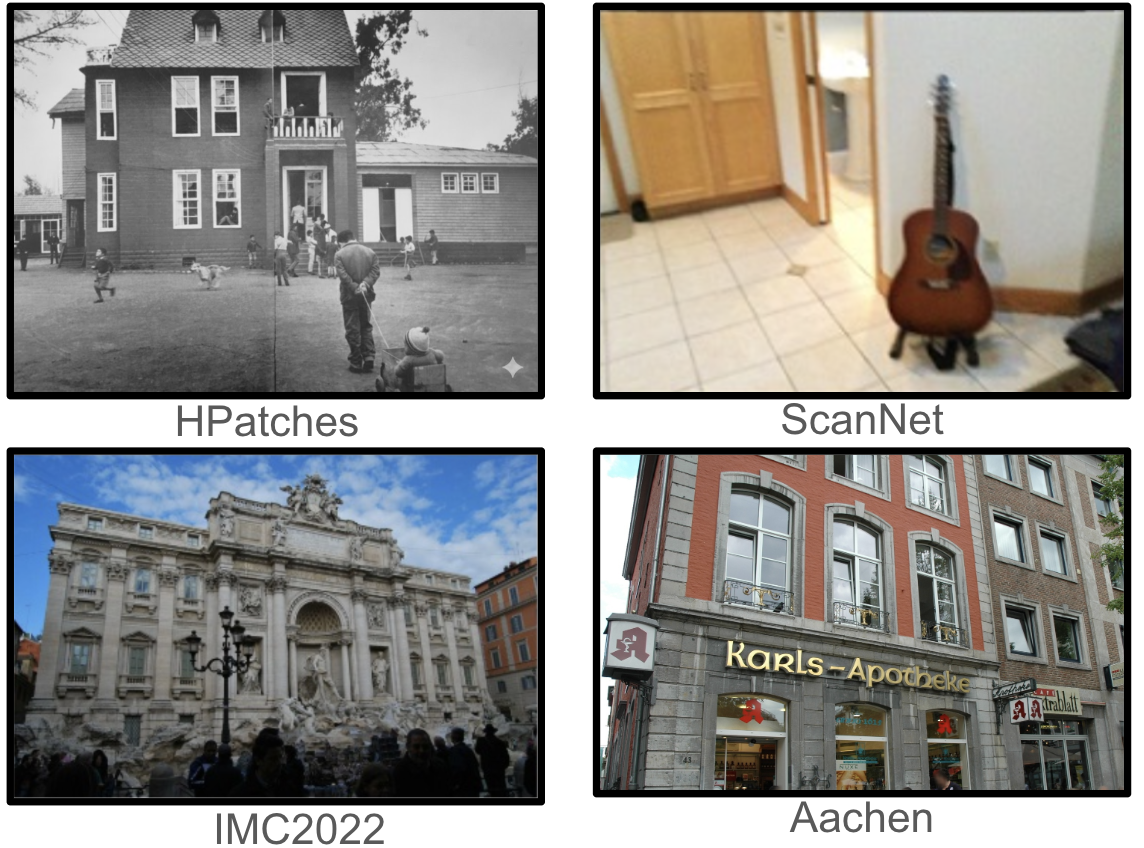}
    \vspace{-3mm}
    \caption{Examples from the evaluation datasets, spanning planar homography scenes (HPatches), indoor environments (ScanNet), and challenging outdoor benchmarks (IMC2022/Aachen).}
    \vspace{-3mm}
    \label{fig:dataset_vis}
\end{figure}

\subsection{Joint Detector $-$ Descriptor Distillation}\label{sec:method_distill}

To further align student and teacher representations beyond explicit correspondences, we introduce a \emph{joint distillation loss} that couples detector confidence and descriptor similarity into a unified probabilistic space. Unlike previous approaches \cite{potje2024xfeat}, which aligns detectors and descriptors independently, our joint formulation models how detector reliability modulates descriptor similarity.

\paragraph{Detector-weighted similarity matrices.}  
Given the raw similarity matrix $\mathbf{S}^{ST}$ and $\mathbf{S}^{TT}$, we define two detector-weighted variants:
% {\small
% \begin{equation}
% \begin{aligned}
% \mathbf{\bar{S}}^{ST}_{ij} &=
% \Big(\tfrac{\sigma_i^S}{\tau_s}\Big)
% \mathbf{S}_{ij}^{ST}
% \Big(\tfrac{\sigma_j^T}{\tau_t}\Big), \\
% \mathbf{\bar{S}}^{TT}_{ij} &=
% \Big(\tfrac{\sigma_i^T}{\tau_t}\Big)
% \mathbf{S}_{ij}^{TT}
% \Big(\tfrac{\sigma_j^T}{\tau_t}\Big).
% \end{aligned}
% \end{equation}
% }

{\small
\begin{equation} \label{ed:joint_dist}
\mathbf{\bar{S}}^{ST}_{ij} =
\Big(\tfrac{\mathbf{w}_i^S}{\tau_s}\Big)
\mathbf{S}_{ij}^{ST}
\Big(\tfrac{\mathbf{w}_j^T}{\tau_t}\Big)
\qquad
\mathbf{\bar{S}}^{TT}_{ij} =
\Big(\tfrac{\mathbf{w}_i^T}{\tau_t}\Big)
\mathbf{S}_{ij}^{TT}
\Big(\tfrac{\mathbf{w}_j^T}{\tau_t}\Big)
\end{equation}
}

% $\mathbf{w}^S$ and $\mathbf{w}^T$

\noindent where $\mathbf{S}_{ij}^{TT}$ denotes the teacher–teacher similarity matrix.
The $\tau_s$ and $\tau_t$ terms are temperatures (selected empirically) controlling the influence of the student and teacher detector confidence. We study their impact in Appendix \ref{app:temperature}.
This produces two joint detector–descriptor spaces:
$\mathbf{\bar{S}}_{ij}^{ST}$ for student–teacher pairs and $\mathbf{\bar{S}}_{ij}^{TT}$for teacher–teacher pairs.

\paragraph{Distillation Loss.}  
Both weighted similarity matrices are converted into probability distributions by applying the previously defined row- and column-wise softmax operators.

The distillation loss is formulated as the sum of row- and column-wise Kullback–Leibler divergences between these distributions:
{\small
\begin{equation}
    \mathcal{L}_{\text{KD}}^{ST}
    =
    \mathrm{KL}\!\big(
    \sigma_r(\mathbf{\bar{S}}^{TT}) \,\|\, \sigma_r(\mathbf{\bar{S}}^{ST})
    \big)
    +
    \mathrm{KL}\!\big(
    \sigma_c(\mathbf{\bar{S}}^{TT}) \,\|\, \sigma_c(\mathbf{\bar{S}}^{ST})
    \big).
\end{equation}
}

Use the similar construction for $\mathcal{L}_{\text{KD}}^{TS}$, the total distillation loss becomes:
% \[
% \mathcal{L}_{\text{KD}} = \mathcal{L}_{\text{KD}}^{ST} + \mathcal{L}_{\text{KD}}^{TS}
% \]

{\small
\begin{equation}
    \mathcal{L}_{\text{KD}} = \mathcal{L}_{\text{KD}}^{ST} + \mathcal{L}_{\text{KD}}^{TS}
\end{equation}
}

\noindent This formulation enforces that the student reproduces the teacher’s joint detector–descriptor distribution along both matching directions, ensuring consistency in both row-wise (query-to-map) and column-wise (map-to-query) similarity structure.

\paragraph{Final objective.}  
The overall AsymLoc loss combines the geometric matching loss with the joint distillation term:

{\small
\begin{equation}\label{eq:total_loss}
    \mathcal{L}_{\text{AsymLoc}}
    =
    \mathcal{L}_{\text{match}}
    +
    \lambda_{\text{KD}}\,
    \mathcal{L}_{\text{KD}},
\end{equation}
}
% \[
% \mathcal{L}_{\text{AsymLoc}}
% =
% \mathcal{L}_{\text{match}}
% +
% \lambda_{\text{KD}}\,
% \mathcal{L}_{\text{KD}},
% \]
where $\lambda_{\text{KD}}$ balances geometric supervision and cross-model probabilistic alignment. This formulation ensures that lightweight student features not only produce geometry-consistent matches but also preserve the teacher’s joint detector–descriptor distribution. Ablation study on $\mathcal{L}_{\text{KD}}$ is available in Appendix \ref{app:temperature}.

\begin{figure*}[t]
    \centering
    \includegraphics[width=\linewidth]{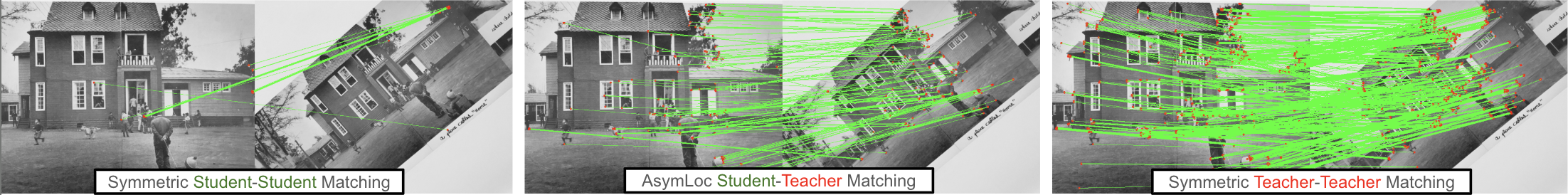}
    \vspace{-6mm}
    \caption{\textbf{AsymLoc student–teacher asymmetric matching visualization.} 
    Symmetric student–student matching fails, whereas asymmetric student–teacher matching succeeds and closely reproduces the teacher–teacher correspondences.}
    \vspace{-2mm}
    \label{fig:match_vis}
\end{figure*}

\section{Experiments}
\label{sec:experiments}

\subsection{Experimental Setup}

\paragraph{Datasets.}  
We evaluate our asymmetric localization framework on four diverse benchmarks, listed below, covering datasets of multiple domains (indoor and outdoor, day/night changes), of multiple tasks (homography estimation, visual localization), and multiple scales (small to large scale); one example per dataset is shown in \Cref{fig:dataset_vis}.
% planar, indoor, and outdoor scenarios.

\noindent \emph{HPatches}~\cite{balntas2017hpatches} provides image pairs with known planar homographies under varying illumination and viewpoint changes. It is primarily used to evaluate homography estimation accuracy and geometric stability of local features.

% \emph{IMC2022}~\cite{imc2022} contains imagery from famous landmarks,
% % is a large-scale outdoor visual localization benchmark created for the Image Matching Challenge at CVPR 2022 Workshop on Image Matching. It
% and measures how accurately query images can be localized within a pre-built reference map. Following the official evaluation protocol, we report mean localization accuracy (MLA) over multiple different  position and orientation thresholds.

% \emph{IMC2022}~\cite{imc2022} contains pairs of outdoor imagery from famous landmarks. We report the mean average accuracy (mAA)~\cite{imc2022} over multiple different position and orientation thresholds. Additional details on the IMC2022 evaluation setup are available in Appendix ~\ref{app:training_details}.

\emph{IMC2022}~\cite{imc2022} consists of outdoor image pairs of famous landmarks. We evaluate our method on random subsets of the training data, averaging the results over multiple seeds. We report the mean average accuracy (mAA)~\cite{imc2022} across multiple position and orientation thresholds.
Additional details on our IMC2022 evaluation setup are available in Appendix~\ref{app:training_details}.

% Mean average accuracy (mAA)

\emph{ScanNet}~\cite{dai2017scannet} consists of scans of indoor environments; 
following standard practice \cite{superglue, loftr, gleize2023silk},
we report the area under the curve (AUC) of pose accuracy at $10^{\circ}$ and $20^{\circ}$ angular thresholds.

\emph{Aachen Day-Night}~\cite{sattler2018aachen} is an outdoor localization dataset with large illumination and appearance changes between day and night. We integrate AsymLoc and the other baselines into the Hierarchical Localization (HLoc) ~\cite{hloc} pipeline, to provide a fair evaluation.

Across IMC2022, ScanNet, and Aachen, we process the database with the teacher and queries with the student; for HPatches, which uses pairs of images, we randomly choose which image is fed to the teacher and which to the student. Additional results on Megadepth \cite{MegaDepthLi18} are reported in \Cref{app:megadepth}.

\paragraph{Implementation Details.}

We train all models using synthetic image pairs generated from the \emph{COCO} dataset~\cite{lin2014coco}.  
Following SiLK~\cite{gleize2023silk}, we sample a single image from COCO and generate a second view by applying a random homographic transformation, yielding a pair $(a,b)$ with known ground-truth homography.  
% This simple homography-based setup has been shown in SiLK to produce strong local feature detectors and descriptors, outperforming alternative training strategies \cite{gleize2023silk}. 
For each training pair $(a,b)$, we use the known homography to obtain ground-truth correspondences $\mathcal{M}_{ab}$. During training, the pre-trained teacher network $T$ remains frozen, while the student network $S$ is optimized using the asymmetric AsymLoc objective discussed above. For the symmetric baselines, both images are encoded by the same network, following standard procedure \cite{superpoint, gleize2023silk}.

Models are trained for $50$ epochs with Adam \cite{KingmaBa2014_adam} with an initial learning rate of $1 \times 10^{-3}$.  
We set the detector confidence threshold $\tau_d$ to $0.65$ and the distillation weight $\lambda_{\text{KD}}$ to $2$ empirically.  
We apply standard data augmentations including random brightness, rotation, scaling, and Gaussian noise.  
To ensure full reproducibility, additional implementation details (learning rate schedule, optimizer settings, hardware setup, temperature ablations, and data augmentation hyperparameters) are provided in Appendix~\ref{app:training_details}.

\begin{table*}[!htbp]\centering
\footnotesize
\resizebox{0.95\textwidth}{!}{%
% \resizebox{\textwidth}{!}{%
\begin{NiceTabular}{l ccccc | cc | cc | c | cc}
\toprule
& \textbf{Method} & \textbf{Asym?} & \makecell{\textbf{\# params} \\ (online \\ model)} & \makecell{\textbf{\# params} \\ (offline \\ model)}  & \makecell{\textbf{GFLOPs}\\(inference on\\1 image)} 
&
% results columns
\multicolumn{2}{c|}{\makecell{\textbf{HPatches} \\ Homography  \\ Est. Accuracy}} &
\multicolumn{2}{c|}{\makecell{\textbf{ScanNet} \\ Relative Pose  \\ Est. AUC}} &
\makecell{\textbf{IMC2022} \\ Mean Avg. \\ Accuracy} &
\multicolumn{2}{c}{\makecell{\textbf{Aachen} \\ Loc. Accuracy \\ (0.5m, $5^\circ$) / (5m, $10^\circ$) }}  
\\

& &&&&& ($\epsilon = 1$) & ($\epsilon = 3$) & \textbf{@$10^{\circ}$} & \textbf{@$20^{\circ}$} &    & Day & Night \\

\cmidrule(l){2-6}\cmidrule(l){7-8}\cmidrule(l){9-10}\cmidrule(l){11-11}\cmidrule(l){12-13} 

% SiLKTeacher ######################################################################################################
\Block[fill=blue!15]{22-1}{\rotatebox{90}{SiLK Teacher}}

% LocSmall
& \Block[fill=gray!13]{1-12}{\textbf{Backbone (0.13M)}} & & & & & & & & & & & \\
& Standard & \xmark & 0.13M &  0.13M & 6.6 & 0.56 & 0.80 & 29.7 & 45.2 & 0.45  & 80.2 / 85.2 & 69.7 / 80.0 \\
& Naive distillation (Symm) & \xmark & 0.13M &  0.13M & 6.6 & 0.56 & 0.79 & 29.5 & 44.2 & 0.44  & 80.0 / 85.1 & 69.9 / 81.0 \\
& Naive distillation (Asym) & \cmark & 0.13M &  1M    & 6.6 & 0.57 & 0.80 & 30.5 & 45.1 & 0.45  & 80.0 / 85.1 & 70.1 / 81.4 \\ 
& AML & \cmark & 0.13M &  1M & 6.6 & 0.57 & 0.81 & 30.7 & 46.2 & 0.46  & 80.6 / 85.5 & 70.1 / 81.4 \\
& RKD & \cmark & 0.13M &  1M & 6.6 & 0.56 & 0.81 & 31.1 & 47.4 & 0.46  & 81.0 / 85.3 & 70.0 / 81.2 \\
& CSD & \cmark & 0.13M &  1M & 6.6 & 0.57 & 0.83 & 32.1 & 47.5 & 0.48  & 81.2 / 85.6 & 71.0 / 82.4 \\
& \Distill & \cmark & 0.13M &  1M & 6.6 & 0.57 & 0.82 & \textbf{32.9} & \textbf{49.0} & 0.47  & 81.5 / 86.0 & 71.0 / 82.4 \\
& AsymLoc (Ours) & \cmark & 0.13M &  1M & 6.6 & \textbf{0.60} & \textbf{0.84} & \textbf{32.9} & 48.9 & \textbf{0.51}  & \textbf{83.3 / 87.8} & \textbf{71.2 / 84.4} \\

% LocMini
& \Block[fill=gray!13]{1-12}{\textbf{ Backbone (0.08M)}} & & & & & & & & & & & \\
& Standard & \xmark & 0.08M &  0.08M & 4.87 & 0.55 & 0.79 & 27.6 & 44.6 & 0.43  & 79.1 / 83.2 & 67.9 / 78.8 \\
% & Naive distillation & \xmark & 0.08M &  0.08M & 4.87 & 0.54 & 0.76 & 26.4 & 44.8 & 0.44  & 79.8 / 83.6 &  67.4 / 78.8 \\
% & AML & \cmark & 0.08M &  1M & 4.87 & 0.56 & 0.81 & 28.3 & 44.9 & 0  & 0/0 & 0/0 \\
% & RKD & \cmark & 0.08M &  1M & 4.87 & 0.55 & 0.80 & 28.6 & 45.1 & 0  & 0/0 & 0/0 \\
% & CSD & \cmark & 0.08M &  1M & 4.87 & 0.56 & 0.83 & 29.5 & 46.1 & 0  & 0/0 & 0/0 \\
% & \Distill & \cmark & 0.08M &  1M & 4.87 & 0.56 & 0.80 & 28.9 & 46.5 & 0  & 0/0 & 0/0 \\
& AsymLoc (Ours) & \cmark & 0.08M &  1M & 4.87 & \textbf{0.59} & \textbf{0.83} & \textbf{31.5 }& \textbf{48.5} & \textbf{0.50}  & \textbf{82.1} / \textbf{86.0} & \textbf{71.0/ 83.2} \\

% LocTiny
& \Block[fill=gray!13]{1-12}{\textbf{ Backbone (0.06M)}} & & & & & & & & & & & \\
& Standard & \xmark & 0.06M &  0.06M & 3.27 & 0.52 & 0.76 & 24.2 & 38.9 & 0.39  & 75.0 / 80.5 & 64.5 / 75.4 \\
% & Naive distillation & \xmark & 0.06M &  0.06M & 3.27 & 0.50 & 0.77 & 24.2 & 39.9 & 0.40  & 75.1 / 79.6 & 64.8 / 75.9 \\
% & AML & \cmark & 0.06M &  1M & 3.27 & 0.53 & 0.78 & 26.7 & 40.9 & 0.42  & 0/0 & 0/0 \\
% & RKD & \cmark & 0.06M &  1M & 3.27 & 0.52 & 0.77 & 25.8 & 42.9 & 0.44  & 0/0 & 0/0 \\
% & CSD & \cmark & 0.06M &  1M & 3.27 & 0.55 & 0.80 & 28.6 & 44.4 & 0.44  & 0/0 & 0/0 \\
% & \Distill & \cmark & 0.06M &  1M & 3.27 & 0.54 & 0.80 & 28.6 & 45.3 & 0.45  & 0/0 & 0/0 \\
& AsymLoc (Ours) & \cmark & 0.06M &  1M & 3.27 & \textbf{0.58} & \textbf{0.83} & \textbf{31.0} & \textbf{47.4} & \textbf{0.48}  & \textbf{80.6 / 85.1} & \textbf{69.2 / 82.0} \\

% LocNano
& \Block[fill=gray!13]{1-12}{\textbf{Backbone (0.04M)}} & & & & & & & & & & & \\
& Standard & \xmark & 0.04M &  0.04M & 1.97 & 0.49 & 0.72 & 22.1 & 35.5 & 0.37  & 73.7 / 78.0 & 60.0 / 73.8 \\
% & Naive distillation & \xmark & 0.04M &  0.04M & 1.97 & 0.50 & 0.74 & 22.5 & 39.5 & 0.39  & 74.1 / 78.3 & 60.2 / 74.0 \\
% & AML & \cmark & 0.04M &  1M & 1.97 & 0.50 & 0.75 & 26.5 & 38.9 & 0  & 0/0 & 0/0 \\
% & RKD & \cmark & 0.04M &  1M & 1.97 & 0.52 & 0.75 & 24.2 & 41.1 & 0  & 0/0 & 0/0 \\
% & CSD & \cmark & 0.04M &  1M & 1.97 & 0.54 & 0.80 & 28.3 & 43.5 & 0  & 0/0 & 0/0 \\
% & \Distill & \cmark & 0.04M &  1M & 1.97 & 0.54 & 0.79 & 27.7 & 43.6 & 0  & 0/0 & 0/0 \\
& AsymLoc (Ours) & \cmark & 0.04M &  1M & 1.97 & \textbf{0.56} & \textbf{0.82} & \textbf{30.1} & \textbf{45.8} & \textbf{0.47}  & \textbf{80.1 }/ \textbf{84.8} & \textbf{69.0 / 81.2} \\

% Oracle
& \Block[fill=gray!13]{1-12}{\textbf{Oracle Teacher Performance}} & & & & & & & & & & & \\
& \textbf{SiLK (Teacher)} & \xmark & 1M &  1M & 47.3 & 0.62 & 0.86 & 34.1 & 50.2 & 0.56  & 87.2 / 91.5 & 74.5 / 86.8\\
\cmidrule(l){2-13} \\

% SuperPoint ######################################################################################################
\Block[fill=orange!15]{9-1}{\rotatebox{90}{SuperPoint Teacher}} 

%  LocMini
& \Block[fill=gray!13]{1-12}{\textbf{Backbone (0.08M)}} & & & & & & & & & & & \\
& Standard & \xmark & 0.08M &  0.08M & 4.87 & 0.38 & 0.74 & 17.5 & 31.0 & 0.38  & 78.5 / 82.1 & 53.0 / 70.2 \\
& AsymLoc (Ours) & \cmark & 0.08M &  1M & 4.87 & \textbf{0.41} & \textbf{0.76} & \textbf{18.3} & \textbf{33.5} & \textbf{0.39}  & \textbf{80.7 / 84.5} & \textbf{56.6 / 72.1}  \\

%  LocTiny
& \Block[fill=gray!13]{1-12}{\textbf{ Backbone (0.06M)}} & & & & & & & & & & & \\
& Standard & \xmark & 0..06M &  0.06M & 3.27 & 0.33 & 0.71 & 12.3 & 26.6 & 0.35  & 73.9 / 78.2 &  51.3 / 68.1 \\
& AsymLoc (Ours) & \cmark & 0.06M &  1M & 3.27& \textbf{0.39} & \textbf{0.75} & \textbf{16.9} & \textbf{31.4} & \textbf{0.37}  & \textbf{77.9 / 83.0} &  \textbf{55.8 / 71.5 }\\

% Oracle
& \Block[fill=gray!13]{1-12}{\textbf{Oracle Teacher Performance}} & & & & & & & & & & & \\
& \textbf{SuperPoint (Teacher)} & \xmark & 1.3M &  1.3M & 26.1 & 0.43 & 0.8 & 21.5 & 36.4 & 0.49  & 86.8 / 90.0 & 59.2 / 74.5 \\
\cmidrule(l){2-13} \\

% % XFeat ######################################################################################################
% \Block[fill=magenta!15]{6-1}{\rotatebox{90}{XFeat Teacher}} 
% %  LocTiny
% & \Block[fill=gray!13]{1-12}{\textbf{ Backbone (0.17M)}} & & & & & & & & & & & \\
% & Standard & \xmark & 0..06M &  0.06M & 3.27 & 0.33 & 0.71 & 12.3 & 26.6 & 0.35  & 73.9 / 78.2 &  51.3 / 68.1 \\
% & AsymLoc (Ours) & \cmark & 0.06M &  1M & 3.27& \textbf{0.39} & \textbf{0.75} & \textbf{16.9} & \textbf{31.4} & \textbf{0.37}  & \textbf{77.9 / 83.0} &  \textbf{55.8 / 71.5 }\\
% % Oracle
% & \Block[fill=gray!13]{1-12}{\textbf{Oracle Teacher Performance}} & & & & & & & & & & & \\
% & \textbf{XFeart (Teacher)} & \xmark & 1.3M &  1.3M & 26.1 & 0.43 & 0.8 & 25.2 & 39.3 & 0.49  & 86.8 / 90.0 & 59.2 / 74.5 \\
% \cmidrule(l){2-13} \\

& \Block[fill=gray!13]{1-12}{\textbf{Reference Models}} & & & & & & & & & & & \\
% & SuperPoint$^\dagger$ & \xmark & 1.3M &  NA & 3.2 & 0 & 0 & 0 & 0 & 0  & 0/0 & 0/0 \\
& LoFTR & \xmark & 28M &  28M & 223 & 0.65 & 0.87 & 40.8 & 57.6 & 0.66  & 94.4 / 97.7 & 91.8 / 98.0 \\
&  SuperPoint + LightGlue & \xmark & 14M &  14M & 63.3 & 0.47 & 0.82 & 35.3 & 53.3 & 0.61  & 95.4 / 98.3 & 91.8 / 100.0 \\

\bottomrule
\end{NiceTabular}
}
    \vspace{-3mm}
    \caption{ 
    \textbf{AsymLoc enables compact student (online) models to achieve localization accuracy competitive with much larger teacher (offline) models.} We present results using [Blue] SiLK and [Orange] SuperPoint as teachers across four diverse datasets: HPatches (homography), ScanNet (indoor), IMC2022 (outdoor), and Aachen (full localization pipeline). By explicitly modeling the asymmetric setup, AsymLoc consistently achieves performance close to the teacher, while standard symmetric settings struggle. Furthermore, AsymLoc outperforms other asymmetric baselines. We report parameters (Params), GFLOPs, and dataset-specific metrics.
    % $\dagger$: Denotes official pre-trained weights.
    Additional ablations are available in Appendix \ref{app:baseline_abl}.
    }
    \vspace{-3mm}
\label{tab:combined}
\end{table*}

\vspace{-1em}
\paragraph{Teacher models.}
AsymLoc can be applied to any model, given that our training pipeline aims at training the student using a pretrained teacher: to showcase this flexibility, we compute experiments with two popular models, namely SiLK \cite{gleize2023silk} and SuperPoint \cite{superpoint}. Additional results using XFeat \cite{potje2024xfeat} are provided in \Cref{app:megadepth}.
% Throughout the experiments, we use \emph{SiLK}~\cite{gleize2023silk} as our teacher network.
% \gb{TODO update with superpoint}
% SiLK provides a clean and lightweight detector–descriptor framework that integrates recent advances in feature distillation, and has been shown to outperform SuperPoint while using fewer parameters. Moreover, SiLK follows a VGG-style backbone with sequential $3{\times}3$ convolutions and ReLU activations, making it a simple and well-controlled base model for evaluating our asymmetric pipeline. Without loss of generality, our approach is not restricted to SiLK and can be applied to any detector–descriptor model.

\paragraph{Student variants.}
We assess the robustness of our training paradigm using four student models with varying capacities, ranging from 0.04M to 0.13M parameters. This design enables a more precise analysis of the size–performance trade-off, and our emphasis on ultra-compact models directly targets edge scenarios such as smart glasses and small-scale mobile robots. Each variant adopts a CNN backbone followed by detector and descriptor heads, mirroring common architectures in the literature (e.g., SuperPoint and SiLK) and thus facilitating direct comparison. Additional experiments, including ResNet-style backbones and models spanning a broader parameter range, are reported in Appendix~\ref{app:architecture_ablations}.

% To prove the robustness of our training paradigm we provide results with four different student models of different sizes, ranging from just 0.04M parameters to 0.13M.
% This allows us to better investigate the size-performance trade-off; moreover this deliberate focus on ultra-small models is critical for edge devices like smart glasses and tiny mobile robotics.
% These variants are implemented with a CNN backbone followed by a detector and a descriptor head: this is similar to common architectures in literature, such as SuperPoint and SiLK, allowing for more direct comparability.
% Additional comparisons, including ResNet-style architectures and models covering a wider parameter range, are provided in Appendix~\ref{app:architecture_ablations}.

The four variant of student architectures are:
\begin{enumerate}
    \item 0.13M parameters / 7-layer CNN backbone.
    \item 0.08M parameters / 7-layer CNN with reduced filters.
    \item 0.06M parameters / 6-layer CNN backbone.
    \item 0.04M parameters / 6-layer CNN with reduced filters.
\end{enumerate}

% Empirically, we find that in the asymmetric setup, models with slightly deeper backbones but narrower channels achieve better trade-offs than those with shallower yet wider networks.

\paragraph{Comparison baselines.}

We compare AsymLoc against the following setups:
\begin{itemize}
    \item \textbf{Oracle (Teacher only):} Both query and database images are processed by the teacher network. This serves as the \emph{oracle} upper bound for accuracy.

    \item \textbf{Standard (Student only):} Both query and map images are processed by the small student model, trained on its own without any teacher supervision. 

    \item \textbf{Naive Distillation:} A standard feature-level distillation baseline in which the student’s descriptor features are trained to directly minimize the cosine distance to the teacher’s corresponding descriptors. The detector logits are supervised using a soft binary cross-entropy (SoftBCE) loss computed on the teacher’s probability maps. We evaluate Naive Distillation is symmetric as well as asymmetric settings.

    \item \textbf{Asymmetric Distillation:} 
    As no previous work tackled the task of asymmetric visual localization, we adapt several methods from the tasks of model distillation and asymmetric image retrieval. Each method supervises the descriptor branch via asymmetric objectives while keeping the detector branch trained using SoftBCE loss:
    \begin{enumerate}
        \item \textbf{Asymmetric Metric Learning (AML)}~\cite{budnik2021aml}: learns a contrastive objective between teacher and student embeddings.
        \item \textbf{Relational Knowledge Distillation (RKD)}~\cite{park2019relational}: aligns pairwise relational distances and angles between samples across teacher and student feature spaces.
        % \mo{Need a suggestion: Both AML and RKD above were proposed as standard distillation techniques but have been applied in an asymmetric manner in CSD and D2Still. Hence I also do the same. Should we mention it?}
        \item \textbf{Contextual Similarity Distillation (CSD)}~\cite{wu2022csd}: distills pairwise similarity scores between teacher features, encouraging the student to maintain the teacher’s similarity structure.
        \item \textbf{Decoupled Differential Distillation (D3Still)}~\cite{xie2024d3still}: extends CSD by additionally transferring pairwise similarity \emph{differentials} to preserve ranking order and relative similarity relationships, and has SOTA performance on asymmetric image retrieval benchmarks.
    \end{enumerate}
\end{itemize}

\subsection{Results}

Table~\ref{tab:combined} presents our main results:
across the four datasets, we present results with SiLK teacher (top part in blue) and SuperPoint teacher (in orange).
We showcase the effect of AsymLoc on these models at different student sizes, providing evaluation metrics, GFLOPS and number of parameters.
Note that for symmetric settings (i.e., Standard and Naive Distillation), the student and teacher models are identical; hence, their parameter counts in the respective columns are the same.
We compare AsymLoc with the aforementioned baselines, as well as a number of popular models for reference, namely SuperPoint+LightGlue and LoFTR, to demonstrate the huge reduction in inference compute brought by AsymLoc.

The results show that AsymLoc nearly closes the gap between tiny models and larger ones, while having the same inference cost as a tiny model:
with the 0.13M student, AsymLoc improves over the \textit{Standard} setup (\ie symmetric tiny models for query and map processing) by 4\%, only 2\% lower than the default SiLK model on HPatches, while being 8 times smaller and requiring 7 times fewer flops.
These results are consistent across every datasets, metrics, model dimension and teacher architecture (both SiLK and SuperPoint); we note in fact that AsymLoc always improves on the \textit{Standard} setup, without any added inference cost. \Cref{fig:match_vis}  shows the AsymLoc matching visualization.

\begin{figure*}[h]
    \centering
    \includegraphics[width=\linewidth]{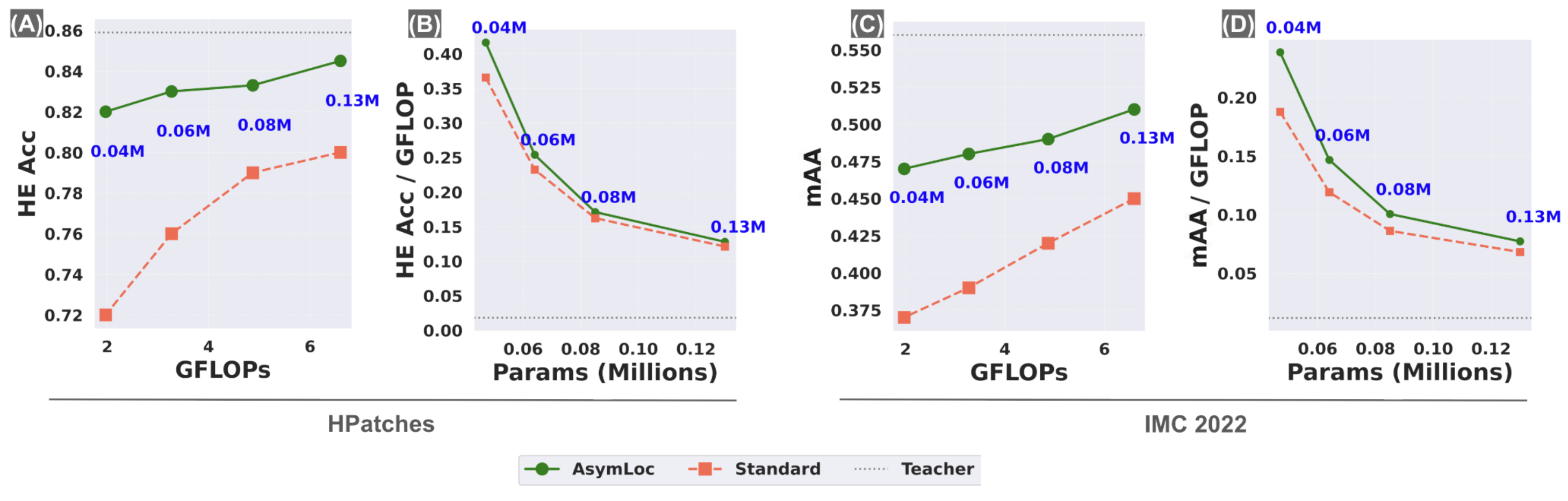}
    \vspace{-4mm}
    \caption{\textbf{Efficiency–accuracy trade-offs for AsymLoc.} 
(A) Homography estimation accuracy (HE Acc) vs. GFLOPs on HPatches. 
(B) HE Acc per GFLOP vs. parameter count. 
(C) Mean average accuracy (mAA) vs. GFLOPs on IMC2022. 
(D) mAA per GFLOP vs. parameter count. 
Across all datasets, asymmetric training yields flatter Pareto curves and higher parameter efficiency, demonstrating superior scalability of AsymLoc compared to standard symmetric training.
}
% \vspace{-4mm}
\label{fig:flops}
\end{figure*}

Across our experiments, we observe that Naive Distillation of a large model into a smaller one provides little to no improvement over the \textit{Standard} setup, proving that using a small model for both query and map leads to lower results regardless of how the small model is trained.
Furthermore, we note that incorporating AML \cite{budnik2021aml} and RKD \cite{park2019relational} leads to consistent gains, indicating that introducing asymmetry between teacher and student representations is beneficial. Significant improvements are achieved with CSD \cite{Asym_CSD_wu2022contextual}, highlighting the importance of distilling similarity structure rather than raw feature values. Unlike in image retrieval, however, adding a ranking loss on top of CSD (following D3Still \cite{Asym_distill_xie2024d3still}) does not yield additional improvements. Finally,
AsymLoc
% by jointly distilling the detector and descriptor representations, our method 
outperforms all existing asymmetric distillation approaches across almost every single metric (with the sole exception of D3still outperforming AsymLoc by 0.1\% on Scannet@20°.

\noindent To further illustrate the trend across asymmetric models, we plot the homography estimation accuracy (HE Acc) against GFLOPs for all models in \Cref{fig:flops}(A). The asymmetric setup exhibits a significantly smaller performance drop rate (in the Pareto curve) compared to standard training. In \Cref{fig:flops}(B), we plot HE Acc per GFLOP against the parameter count to highlight parameter efficiency. As expected, all models become more parameter-efficient as the number of parameters decreases—a common trend in most machine learning setups, since additional parameters yield diminishing returns. However, the efficiency of AsymLoc improves at a much faster rate than that of the standard models, as clearly visible in the trend. 
We observe similar results on the IMC2022 dataset, where we plot the mean average accuracy (mAA) against GFLOPs in \Cref{fig:flops}(C), and mAA per GFLOP against the parameter count in \Cref{fig:flops}(D). We report additional latency analysis in \Cref{app:speed_vs_acc}.

These results collectively demonstrate that AsymLoc provides a general solution to edge-device localization: lightweight query models remain fully compatible with heavy teacher-derived map features, achieving near-teacher performance at a fraction of the compute and memory cost.

\subsection{Ablation}
We conducted an ablation study to analyze the impact of our two loss components, $\mathcal{L}_{\text{match}}$ and $\mathcal{L}_{\text{KD}}$, with results presented in Table~\ref{tab:ablation_modified}. The analysis reveals that $\mathcal{L}_{\text{match}}$, when applied in isolation, is detrimental to performance. This is because $\mathcal{L}_{\text{match}}$ lacks a negative signal for the detector; it functions primarily as a regularizer that re-weights the loss to prioritize regions where the teacher model is confident. Conversely, $\mathcal{L}_{\text{KD}}$ alone provides a significant performance boost. The optimal result is achieved by combining both terms, which yields a further improvement and indicates a synergistic relationship between the two components.

\begin{table}
\footnotesize
\begin{center}
\begin{adjustbox}{width=0.9\columnwidth}
\begin{tabular}{cc|cc c cc}
\toprule
$\mathcal{L}_{\text{match}}$ & $\mathcal{L}_{\text{KD}}$
& \multicolumn{2}{c}{\textbf{HPatches}} && \multicolumn{2}{c}{ \textbf{ScanNet}} \\
\cline{3-4} \cline{6-7}

& & \multicolumn{2}{c}{ HEA} && \multicolumn{2}{c}{ RP-AUC} \\
\cline{3-4} \cline{6-7}
& & ($\epsilon = 1$)  & ($\epsilon = 3$)  &&  \textbf{@$10^{\circ}$} & \textbf{@$20^{\circ}$}\\
\midrule
\checkmark & & 0.53 & 0.70 && 21.6  &	35.8 \\
& \checkmark &  0.57&	0.82 &&	30.0 &	46.9 \\
\checkmark & \checkmark & \textbf{0.59} & \textbf{0.83} && \textbf{31.5} & \textbf{48.5} \\
\bottomrule
\end{tabular}
\end{adjustbox}
\end{center}
\vspace{-4mm}
\caption{
\small
Analyzing the impact of $\mathcal{L}_{\text{match}}$ and $\mathcal{L}_{\text{KD}}$ on HPatches and ScanNet Datasets. We report Homography Estimation Accuracy (HEA) for HPatches and Relative Pose Prediction AUC (RP-AUC) for ScanNet. 
}
\vspace{-3mm}
\label{tab:ablation_modified}
\end{table}

\section{Conclusion}
\label{sec:conclusion}

% \mo{This para is chatgpt. Looks okay?}
We introduced AsymLoc, a visual localization framework that, despite incurring in the same inference cost of tiny models, achieves similar results as standard bigger models.
AsymLoc attains this by being the first visual localization pipeline that relies on two different models for processing the database (performed offline) and the queries (online, on-device).
To align the two models, we overcame the limitations of existing baselines with a novel distillation objective that aligns models in the \textit{joint detector-descriptor} space, combining a geometric matching loss with a probabilistic alignment of feature interactions.
This approach allows ultra-lightweight student models (as small as 0.04M parameters) to be directly compatible with 1.0M parameter teachers. Across diverse planar, indoor, and large-scale outdoor benchmarks, our 25$\times$ smaller student models retain over 96\% of the teacher's accuracy, decisively outperforming symmetric baselines and prior asymmetric distillation methods, paving the way for visual localization frameworks that can efficiently run on edge devices with massive reduction of inference cost.

\section{Acknowledgment}
We thank Amazon for their support during the summer internship and through the Amazon AI PhD Fellowship.

% \newpage
{
    \small
    \bibliographystyle{ieeenat_fullname}
    \bibliography{main}
}
\newpage

\appendix

\section{Appendix}
\subsection{Hyperparameter Ablations}\label{app:temperature}
In equation \ref{ed:joint_dist}, we defined detector-weighted similarity matrices as:

{\small
$
\mathbf{\bar{S}}^{ST}_{ij} =
\Big(\tfrac{\mathbf{w}_i^S}{\tau_s}\Big)
\mathbf{S}_{ij}^{ST}
\Big(\tfrac{\mathbf{w}_j^T}{\tau_t}\Big)
\qquad
\mathbf{\bar{S}}^{TT}_{ij} =
\Big(\tfrac{\mathbf{w}_i^T}{\tau_t}\Big)
\mathbf{S}_{ij}^{TT}
\Big(\tfrac{\mathbf{w}_j^T}{\tau_t}\Big)
$
}

% \emph{\noindent \textbf{Note:} We use the notations $\mathbf{\sigma}^S$ and $\mathbf{\sigma}^T$ for the student and teacher detector confidences \textbf{interchangeably} with $\mathbf{w}^S$ and $\mathbf{w}^T$, respectively, throughout this paper.}

\begin{table}[h]
\begin{adjustbox}{width=\columnwidth}
\tiny
\begin{tabular}{c c c c | c c c c } % Defining 8 columns, with a vertical line after column 4
\toprule
\multicolumn{4}{c|}{$\mathbf{\tau_s}$} & 1 & 0.5 & 0.1 & 0.5 \\
\multicolumn{4}{c|}{$\mathbf{\tau_t}$} & 1 & 0.5 & 0.1 & 0.1 \\
\midrule
\multicolumn{3}{c}{\multirow{2}{*}{\textbf{HPatches}}} & $\epsilon = 1$ & 0.59 & \textbf{0.60} & 0.57 & 0.58 \\
% \cline{4-8} % Draw a line only under the data columns (columns 4 to 8)
\multicolumn{3}{c}{} &                                   $\epsilon = 3$ & 0.82 & \textbf{0.84} & 082 & 0.81 \\
\bottomrule
\end{tabular}
\end{adjustbox}
\caption{Ablation study of the temperature parameters $\mathbf{\tau_s}$ and $\mathbf{\tau_t}$ used in the Joint Detector--Descriptor Distillation loss.}
\label{tab:app_temp}
\end{table}

This formulation allows us to control the influence of the student and teacher detector confidences in the joint similarity space $\mathbf{\bar{S}}$.
Table \ref{tab:app_temp} investigates the impact of the temperature parameters $\mathbf{\tau_s}$ and $\mathbf{\tau_t}$ on performance. We observe that decreasing the temperature values (which corresponds to increasing the influence of the detector confidences) from $1.0$ to $0.5$ generally improves performance. However, performance begins to decline when the temperature is decreased further (e.g., below $0.5$). Furthermore, the use of different temperature values for the student ($\mathbf{\tau_s}$) and teacher ($\mathbf{\tau_t}$) confidences does not appear to offer any significant benefit.

\begin{table}
\begin{center}
\begin{adjustbox}{width=\columnwidth}
\tiny
\begin{tabular}{c|cc c cc}
\toprule
$\lambda_{\text{KD}}$
& \multicolumn{2}{c}{\textbf{HPatches}} && \multicolumn{2}{c}{ \textbf{ScanNet}} \\
& \multicolumn{2}{c}{ HEA} && \multicolumn{2}{c}{ RP-AUC} \\
 & ($\epsilon = 1$)  & ($\epsilon = 3$)  &&  \textbf{@$10^{\circ}$} & \textbf{@$20^{\circ}$}\\
\midrule
0 & 0.53 & 0.70 && 21.6  &	35.8 \\
1 &  0.54 &	0.79 &&	29.5 &	45.0 \\
2 & \textbf{0.59} & \textbf{0.83} && \textbf{31.5} & \textbf{48.5} \\
4 &  0.57&	0.81 &&	30.0 &	47.0 \\
\bottomrule
\end{tabular}
\end{adjustbox}
\end{center}
\caption{Analyzing the impact of $\lambda_{\text{KD}}$ on HPatches and ScanNet Datasets. We report Homography Estimation Accuracy (HEA) for HPatches and Relative Pose Prediction AUC (RP-AUC) for ScanNet. }
\label{tab:app_lkd}
\end{table}

In Equation \ref{eq:total_loss}, we defined our total training loss as a weighted combination of the geometric matching loss ($\mathcal{L}_{\text{match}}$) and the joint detector--descriptor distillation loss ($\mathcal{L}_{\text{KD}}$), balanced by the factor $\lambda_{\text{KD}}$. Table \ref{tab:app_lkd} presents an ablation study on the impact of this weighing factor. When $\mathbf{\lambda_{\text{KD}} = 0}$, only the matching loss is utilized, yielding results identical to the $\mathcal{L}_{\text{match}}$-only case reported in Table \ref{tab:ablation_modified}. As $\mathbf{\lambda_{\text{KD}}}$ is increased, performance steadily improves, reaching maximum performance at $\mathbf{\lambda_{\text{KD}} = 2}$. Further increasing $\lambda_{\text{KD}}$ shifts the balance toward the distillation loss, leading to results that approach those of the $\mathcal{L}_{\text{KD}}$-only case presented in Table \ref{tab:ablation_modified}.

\subsection{Exploring Different Architectures}\label{app:architecture_ablations}

\begin{figure}[h]
    \centering
    \includegraphics[width=\linewidth]{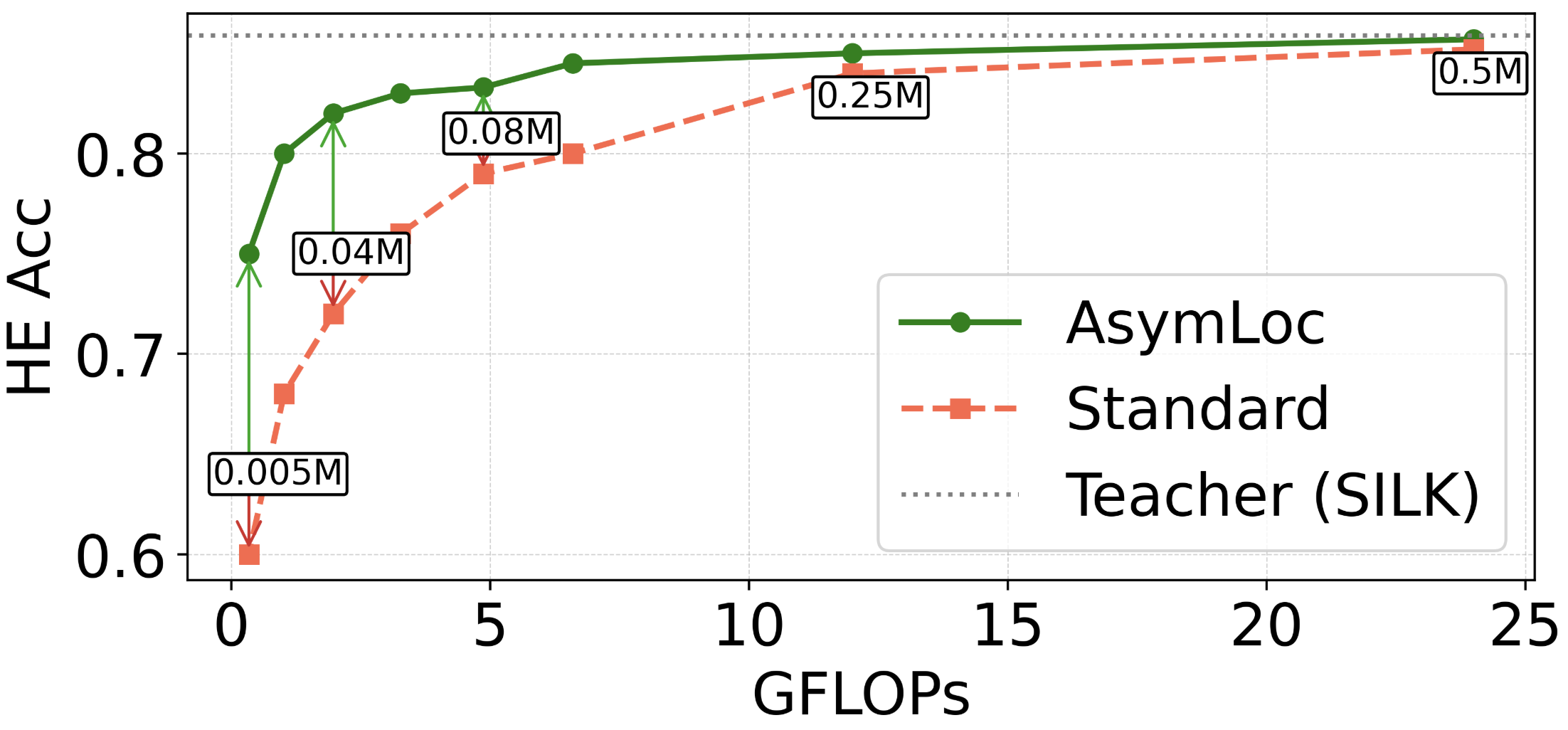}
    \caption{Homography estimation accuracy on HPatches with a wide range of model sizes. Here we use SILK as the teacher.}
    \label{fig:app_more_sizes}
\end{figure}

While the main paper presented results for four model sizes ($0.12 \text{M}$, $0.08 \text{M}$, $0.06 \text{M}$, and $0.04 \text{M}$ parameters), Figure \ref{fig:app_more_sizes} provides an extended analysis incorporating a wider range of model complexities, including much smaller ($0.02 \text{M}$ and $0.005 \text{M}$) and larger ($0.25 \text{M}$ and $0.5 \text{M}$) models. We observe that for model sizes of $\mathbf{0.25 \text{M}}$ parameters and above, the performance of both the symmetric and Asymmetric pipelines closely approximates that of the full Teacher model, with the Asymmetric approach exhibiting a slight advantage. As the parameter count is reduced, the Asymmetric pipeline proves significantly more robust, retaining performance much better than its symmetric counterpart. Specifically, the Asymmetric approach maintains performance close to the Teacher's down to $\mathbf{0.04 \text{M}}$ parameters, but then begins to show a sharp decline at $\mathbf{0.02 \text{M}}$ parameters, though it still outperforms the symmetric pipeline.

Table \ref{tab:app_resnet} presents an analysis of the impact of incorporating  residual connections into our pipeline. We find that the addition of these connections offers no major performance advantage, likely due to the fact that our pipeline utilizes only small CNN, which do not typically suffer from the vanishing gradient issues that residual connections are designed to mitigate in deeper architectures.

\begin{table}
\begin{center}
\begin{adjustbox}{width=\columnwidth}
\footnotesize
\begin{tabular}{c|cc c cc}
\toprule
\textbf{Param} & \multicolumn{2}{c}{\textbf{HPatches}} && \multicolumn{2}{c}{ \textbf{HPatches}} \\ \\
\makecell{Residual?} & \multicolumn{2}{c}{ } && \multicolumn{2}{c}{ \checkmark} \\
 & ($\epsilon = 1$)  & ($\epsilon = 3$)  & & ($\epsilon = 1$)  & ($\epsilon = 3$) \\
\midrule
0.13M & 0.60 & 0.84 && 0.60  &	0.83 \\
0.8M &  0.59 &	0.83 &&	0.59 &	0.83 \\
\bottomrule
\end{tabular}
\end{adjustbox}
\end{center}
\vspace{-1em}
\caption{Analyzing the impact of adding residual connections. }
\vspace{-1em}

\label{tab:app_resnet}
\end{table}

\subsection{Training and Evaluation Details}\label{app:training_details}
Our pipeline was trained using the $\mathbf{Adam}$ optimizer with a base learning rate of $\mathbf{0.001}$ and standard momentum settings ($\mathbf{\beta_1 = 0.9}, \mathbf{\beta_2 = 0.999}$). To enhance robustness, we employed a comprehensive suite of data augmentation techniques (adopted from \cite{gleize2023silk}). This suite included color and exposure manipulations such as $\text{Random Gamma}$ and $\text{Hue, Saturation, and Value}$ shifts, various blurring effects including standard $\text{Blur}$ and $\text{Motion Blur}$, and $\text{Gaussian Noise}$ injection. Furthermore, we applied $\text{Random Brightness and Contrast}$ adjustments to broaden the model's tolerance to varying lighting conditions.

For our IMC2022 evaluations, we use random subsets of the training set with different seeds and average the results. Note that none of our models are trained on the IMC2022 training set. Furthermore, this setting differs from the official IMC2022 benchmark, which requires submissions on Kaggle—a process that is prohibitively time-consuming considering the number of models we are evaluating.

\subsection{Additional Baseline Ablations}\label{app:baseline_abl}
Table \ref{tab:combined} presents our main results, demonstrating that asymmetric matching outperforms standard symmetric matching. We further compare our method against other asymmetric baselines, showing that our formulation yields superior performance. Due to space constraints, the main analysis focuses on the 0.13M parameter models. For completeness, we report results for additional model capacities (0.08M and 0.06M) on HPatches and ScanNet in Tables \ref{tab:hpatches_truncated} and \ref{tab:scannet_truncated}.

\begin{table}[!htbp]\centering
\footnotesize
\begin{tabular}{cccccccc}
\toprule
&  & &\multicolumn{2}{c}{\textbf{(0.08M)}} & &\multicolumn{2}{c}{\textbf{ (0.06M)}} \\
\cmidrule{4-5}\cmidrule{7-8}
\textbf{Asym?} &\textbf{Technique} &  &($\epsilon = 1$) &($\epsilon = 3$) &  &($\epsilon = 1$) &($\epsilon = 3$) \\
\midrule
\xmark & \makecell{Standard}            & & 0.55 & 0.79 & & 0.52 & 0.76\\
\xmark &\makecell{Naive \\ Distillation}& & 0.54 & 0.76 & & 0.50 & 0.77\\
\cmark &AML                             & & 0.56 & 0.81 & & 0.53 & 0.78\\
\cmark &RKD                             & & 0.55 & 0.80 & & 0.52 & 0.77\\
\cmark &CSD                             & & 0.56 & \textbf{0.83} & & 0.55 & 0.80\\
\cmark &\Distill{}                      & & 0.56 & 0.80 & & 0.54 & 0.80\\
\cmark &Ours                            & &\textbf{0.59} & \textbf{0.83}& &\textbf{0.58} & \textbf{0.83}\\
\bottomrule
\end{tabular}
\caption{\textbf{Homography estimation accuracy on HPatches (0.08M and 0.06M).}}
\label{tab:hpatches_truncated}
\end{table}

\begin{table}[!htbp]\centering
\footnotesize
\begin{tabular}{cccccccc}
\toprule
&  & &\multicolumn{2}{c}{\textbf{(0.08M)}} & &\multicolumn{2}{c}{\textbf{ (0.06M)}} \\
\cmidrule{4-5}\cmidrule{7-8}
\textbf{Asym?} &\textbf{Technique} &  &\textbf{@$10^{\circ}$} &\textbf{@$20^{\circ}$} & &\textbf{@$10^{\circ}$} &\textbf{@$20^{\circ}$} \\
\midrule
\xmark & \makecell{Standard}            & & 27.6 & 44.6 & & 24.2 & 38.9 \\
\xmark &\makecell{Naive \\ Distillation}& & 26.3 & 44.8 & & 24.2 & 39.9 \\
\cmark &AML                             & & 28.3 & 44.9 & & 26.7 & 40.9 \\
\cmark &RKD                             & & 28.6 & 45.1 & & 25.8 & 42.9 \\
\cmark &CSD                             & & 29.5 & 46.1 & & 28.6 & 44.4 \\
\cmark &\Distill{}                      & & 28.9 & 46.5 & & 28.6 & 45.3 \\
\cmark &Ours                            & &\textbf{31.5} & \textbf{48.5}& &\textbf{31.0}& \textbf{47.4} \\
\bottomrule
\end{tabular}
\caption{\textbf{Relative pose estimation accuracy on ScanNet (0.08M and 0.06M).}}
\label{tab:scannet_truncated}
\end{table}

\subsection{Speed vs Accuracy}\label{app:speed_vs_acc}

\begin{figure}[!h]
    % \small
    \centering
    \includegraphics[width=\linewidth]{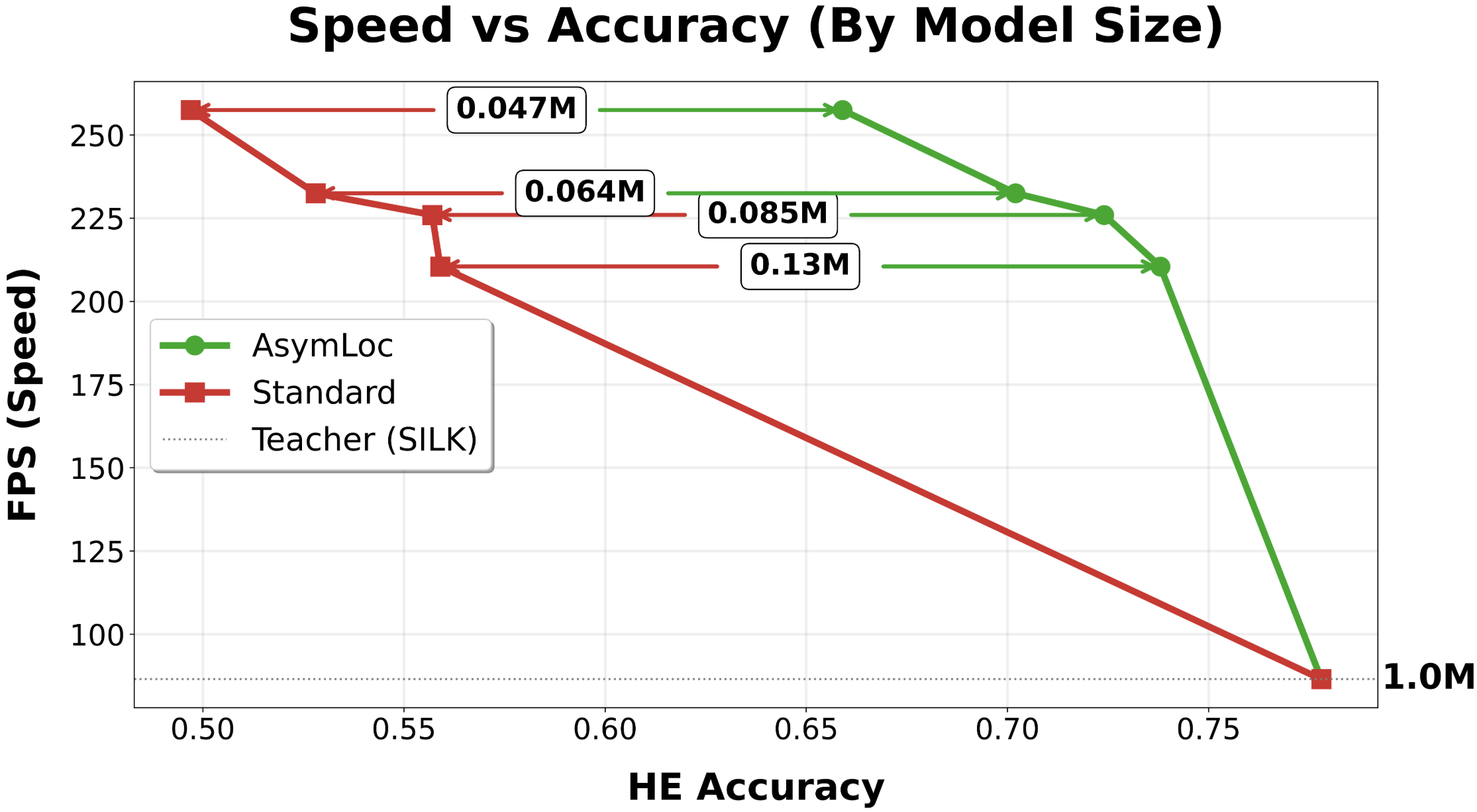}
    \caption{FPS Comparison on HPatches with SILK Teacher}
    \label{fig:fps}
\end{figure}

We report additional latency analysis in \Cref{fig:fps}, plotting FPS vs. Homography Estimation Accuracy on HPatches using SILK as the teacher model. This analysis is done on an NVIDIA RTX A5000 GPU.

\subsection{Additional Results with XFeat}\label{app:megadepth}

XFeat is trained on both COCO homography (like us) as well as MegaDepth-v1. For thoroughness, we evaluate two asymmetric configurations: (1)
\textbf{ XFeat-Mega}, using the official pre-trained weights as the teacher and training the student models on both COCO and MegaDepthv1; (2)
\textbf{XFeat-COCO}, training XFeat teacher and students solely on COCO homography.
We show results on both ScanNet as well as MegaDepth-1500.
Results indicate that the asymmetry consistently outperforms the symmetric baseline by a wide margin, and that asymmetry nears the oracle teacher performance.
\begin{table}[h]\centering
% \footnotesize
\resizebox{\linewidth}{!}{%
\begin{NiceTabular}{l cccc | cc | cc}
\toprule
& \textbf{Method} & \textbf{Asym?} & \makecell{\textbf{\# params} \\ (online)} & \makecell{\textbf{\# params} \\ (offline)}
& \multicolumn{2}{c|}{\makecell{\textbf{ScanNet} \\ Rel. Pose AUC}} & \multicolumn{2}{c}{\makecell{\textbf{MegaDepth-1500} \\ Rel. Pose AUC}} \\
& &&&&   \textbf{@10$^{\circ}$} & \textbf{@20$^{\circ}$} & \textbf{@10$^{\circ}$} & \textbf{@20$^{\circ}$} \\
\cmidrule(l){2-5}\cmidrule(l){6-7}\cmidrule(l){8-9}

\Block[fill=orange!15]{5-1}{\rotatebox{90}{XFeat (Mega)}}
& \Block[fill=gray!13]{1-8}{\textbf{Backbone (0.17M)}} & & & & & & &   \\
& Standard & \xmark & 0.17M & 0.17M & 25.2 & 39.3 & 66.5 & 72.4  \\
& AsymLoc (Ours) & \cmark & 0.17M & 1.5M & 30.2  &  44.4 & 71.3 & 78.6   \\
& \Block[fill=gray!13]{1-8}{\textbf{Oracle Teacher Performance}} & & & &  \\
& \makecell{\textbf{XFeat (Teacher)} \\ (Pre-trained)} & \xmark & 1.5M & 1.5M & 32.2  &  46.8 & 75.2 & 81.9   \\

% \\
\hline

\Block[fill=blue!15]{6-1}{\rotatebox{90}{XFeat (COCO)}}
& \Block[fill=gray!13]{1-8}{\textbf{Backbone (0.17M)}} & & & & & & &   \\
\\
& Standard & \xmark & 0.17M & 0.17M & 24.0 & 38.5 & 53.7& 65.9  \\
& AsymLoc (Ours) & \cmark& 0.17M & 1.5M & 29.2 & 43.2 & 60.1 & 71.1  \\
& \Block[fill=gray!13]{1-8}{\textbf{Oracle Teacher Performance}} & & & & & & &   \\
& \textbf{XFeat (Teacher)} & \xmark  & 1.5M & 1.5M & 31.3 & 46.6 & 65.1 & 73.7   \\

\bottomrule
\end{NiceTabular}
}
\caption{Results with XFeat on ScanNet and MegaDepth}
\label{tab:suppl}
\end{table}


\end{document}